\documentclass{article}
\usepackage[linesnumbered,ruled,vlined]{algorithm2e}
\usepackage{spconf,amsmath,graphicx,algpseudocode}
\usepackage{multirow,diagbox,booktabs}

\usepackage{color}

\title{LBF: LEARNABLE BILATERAL FILTER FOR POINT CLOUD DENOISING}
\name{Huajian Si$^1$, Zeyong Wei$^1$, Zhe Zhu$^1$, Honghua Chen$^1$, Dong Liang$^1$, Weiming Wang$^2$, Mingqiang Wei$^1$}
\address{$^1$Nanjing University of Aeronautics and Astronautics \ \ $^2$Hong Kong Metropolitan University}
\begin{document}
%
\maketitle
\begin{abstract}

Bilateral filter (BF) is a fast, lightweight and effective tool for image denoising and well extended to point cloud denoising. However, it often involves continual yet manual parameter adjustment; this inconvenience discounts the efficiency and user experience to obtain satisfied denoising results. 
We propose \textbf{LBF}, an end-to-end learnable bilateral filtering network for point cloud denoising; \textit{to our knowledge, this is the first time.} 
Unlike the conventional BF and its variants that receive the same parameters for a whole point cloud, LBF learns adaptive parameters for each point according its geometric characteristic (e.g., corner, edge, plane), avoiding remnant noise, wrongly-removed geometric details, and distorted shapes.
Besides the learnable paradigm of BF, we have two cores to facilitate LBF. First, different from the local BF, LBF possesses a global-scale feature perception ability by exploiting multi-scale patches of each point. Second, LBF formulates a geometry-aware bi-directional projection loss, leading the denoising results to being faithful to their underlying surfaces. Users can apply our LBF without any laborious parameter tuning to achieve the optimal denoising results.  
Experiments show clear improvements of LBF over its competitors on both synthetic and real-scanned datasets.

\end{abstract}
\section{Introduction}
\label{sec:intro}
With the popularity of 3D scanners, point clouds are widely used in research fields such as autonomous driving and 3D reconstruction. However, point clouds obtained from consumer-level 3D scanners are often noisy due to equipment measurement precision, object surface textures, etc. Noise seriously affects its characteristics and accuracy, making point cloud denoising a significant preprocessing step.

Existing point cloud denoising methods can be divided into traditional geometric approaches \cite{BF, LOP, WLOP, CLOP, L0, EAR, MLS} and deep learning-based approaches \cite{PCN, EC-Net, TD, PF, Geodualcnn, Repcd, MODNet}. Among them, BF is one of the most widely-used denoising methods. It was first presented by Tomasi and Manduchi \cite{BF-Image} for image denoising and well adapted to 3D mesh denoising by Fleishman et al. \cite{BF-mesh}. Following the experience of BF in mesh denoising, Digne et al. \cite{BF} extend it to point cloud denoising for the first time. Although BF has achieved satisfactory denoising performance, it still has some defects. 
When dealing with point clouds of different point densities and noise intensities, BF requires continual manual tuning of parameters, but finding the best parameters is non-trivial. 
Besides, BF and its variants take the same parameters for all points in the whole point cloud, which leads to under- or over-smoothing of data with rich geometric structures (corners, smooth features, and sharp edges) and multiple noise intensities.
Thus, it is inappropriate to process all points of a noisy point cloud with the same parameters.

To solve the above problems, we propose an end-to-end learnable bilateral filtering network for point cloud denoising, called LBF.
Different from BF, LBF learns adaptive parameters for each noisy point based on its geometric characteristic. In this way, noise in flattened areas is effectively removed, while geometric details in sharp areas can be well preserved. 
Besides the learnable paradigm of BF, we propose two techniques to facilitate LBF. First, unlike the local BF, LBF takes multi-scale patches of each point as input to achieve  global feature perception.
Second, we design a geometry-aware bi-directional projection loss function, enabling LBF to denoise in a feature-aware manner.
Extensive experiments demonstrate that our LBF outperforms the state-of-the-art methods in terms of quantitative results and visual quality.

\begin{figure*}
    \centering
    \includegraphics[width=0.98\textwidth]{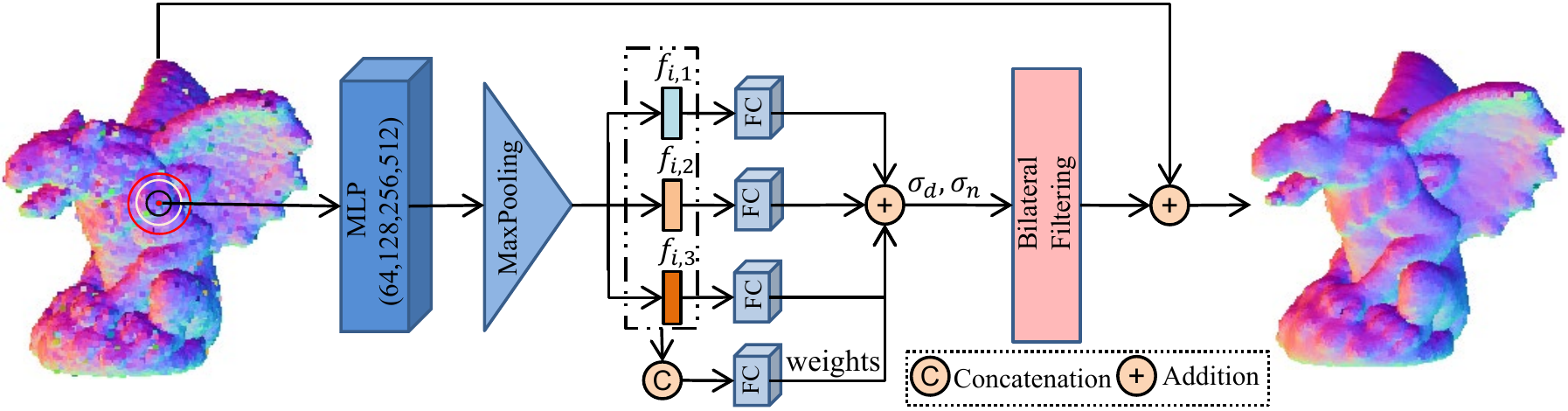}
    \caption{The overview of learnable bilateral filter (LBF).}
    \label{fig:overview}
\end{figure*}

\section{METHOD}
\label{sec:method}
\subsection{Problem Formulation and Overview}
\label{sub2.2}
Given a noisy point cloud, we aim to obtain the noise-free version by removing the additive noise. We first formulate the clean point cloud intuitively as:
\begin{equation}
    \footnotesize
    \mathcal{P}=\hat{\mathcal{P}}+\mathcal{D},
\end{equation} 
where $\hat{\mathcal{P}}=\left\{\hat{p_1}, \hat{p_2}, \ldots, \hat{p_n} \mid \hat{p_n}\in R^3\right\}$ is an observed noisy point cloud, $\mathcal{P}$ is the corresponding clean point cloud and $\mathcal{D}$ is the displacement between the noisy point and noise-free point. Existing point cloud denoising approaches always calculate this displacement and then add it back to the noisy point coordinates to obtain the denoised results. 

Classical BF computes the displacement $\mathcal{D}$ in two steps. First, a unit unoriented normal $n_p$ is computed for each point $p$ according to the principal component analysis for its neighbors $\mathcal{N}_r(p)=$ $\left\{q \in \hat{\mathcal{P}} \mid\|q-p\|_2<r\right\}$. Then, the displacement is calculated by $\mathcal{D}=\delta p \cdot n_p$, and $\delta p$ is defined as
\begin{equation}
    \footnotesize
    \delta p=\frac{\sum_{q \in \mathcal{N}_r(p)} w_d(\|q-p\|)
    w_n\left(\left|\left\langle n_p,
    q-p\right\rangle\right|\right)\left\langle n_p, q-p\right\rangle}{\sum_{q \in
    \mathcal{N}_r(p)} w_d(\|q-p\|) w_n\left(\left|\left\langle n_p,
    q-p\right\rangle\right|\right)},
\end{equation}
where $w_d$ and $w_n$ are two Gaussian weights whose parameters $\sigma_d$ and $\sigma_n$ are manually adjusted by users.
Since these two parameters are difficult to adjust, to address these issues, we propose to use a neural network, to learn the optimal parameters ($\sigma_d$ and $\sigma_n$) for each noisy point, which can produce geometry-adaptive displacement $\delta p$. Consequently, the final denoising model is re-formulated as:
\begin{equation}
    \footnotesize
    \mathcal{P}=\hat{\mathcal{P}} + BF(f(\hat{\wp}_{i, k})),
\end{equation} 
where $BF(\cdot)$ means bilateral filtering, $\hat{\wp}_{i, k}$ is the multi-scale neighborhood of each point, $f(\cdot)$ is a neural network that takes local point patches as input to learn the parameters.
Fig. \ref{fig:overview} shows how LBF to work. LBF is an encoder-decoder network to learn adaptive parameters for each noisy point. The encoder module extracts global geometric features from multi-scale local patches. Then, the multi-scale features are fed into a decoder module to predict $\sigma_d$ and $\sigma_n$, which are further fed into the bilateral filter to obtain the denoised result. We also present an end-to-end training scheme with a geometry-aware bi-directional projection loss function, so that the parameter prediction network can enable more sharp features to be considered during denoising.
\begin{table*}[t]
    \small
    \centering
    \begin{tabular}{|c|c|ccc|ccc|ccc|c|}
    \hline
    \multirow{2}{*}{Metrics} & \multirow{2}{*}{Methods} & \multicolumn{3}{c|}{0.5\%} & \multicolumn{3}{c|}{1.0\%} & \multicolumn{3}{c|}{1.5\%} & \multirow{2}{*}{Average} \\ \cline{3-11}
     &  & \multicolumn{1}{c|}{10k} & \multicolumn{1}{c|}{20k} & 50k & \multicolumn{1}{c|}{10k} & \multicolumn{1}{c|}{20k} & 50k & \multicolumn{1}{c|}{10k} & \multicolumn{1}{c|}{20k} & 50k &  \\ \hline
    \multirow{6}{*}{CD ($10^{-5}$)} & BF & \multicolumn{1}{c|}{4.31} & \multicolumn{1}{c|}{4.08} & 3.33 & \multicolumn{1}{c|}{6.26} & \multicolumn{1}{c|}{4.3} & 3.27 & \multicolumn{1}{c|}{7.68} & \multicolumn{1}{c|}{5.44} & 3.97 & 4.74 \\ \cline{2-12} 
     & WLOP & \multicolumn{1}{c|}{6.28} & \multicolumn{1}{c|}{5.43} & 4.26 & \multicolumn{1}{c|}{9.14} & \multicolumn{1}{c|}{7.55} & 7.07 & \multicolumn{1}{c|}{12.32} & \multicolumn{1}{c|}{9.71} & 7.84 & 7.73 \\ \cline{2-12} 
     & EC-Net & \multicolumn{1}{c|}{5.24} & \multicolumn{1}{c|}{2.85} & 1.30 & \multicolumn{1}{c|}{6.41} & \multicolumn{1}{c|}{3.74} & 2.08 & \multicolumn{1}{c|}{7.90} & \multicolumn{1}{c|}{5.14} & 4.11 & 4.31 \\ \cline{2-12} 
     & PF & \multicolumn{1}{c|}{3.13} & \multicolumn{1}{c|}{2.05} & 1.04 & \multicolumn{1}{c|}{5.87} & \multicolumn{1}{c|}{3.31} & 1.59 & \multicolumn{1}{c|}{8.51} & \multicolumn{1}{c|}{4.79} & 2.42 & 3.63 \\ \cline{2-12} 
     & RePCD & \multicolumn{1}{c|}{3.02} & \multicolumn{1}{c|}{2.08} & 1.02 & \multicolumn{1}{c|}{5.42} & \multicolumn{1}{c|}{3.13} & \textbf{1.57} & \multicolumn{1}{c|}{6.96} & \multicolumn{1}{c|}{4.40} & \textbf{2.38} & 3.33 \\ \cline{2-12} 
     & Ours & \multicolumn{1}{c|}{\textbf{2.87}} & \multicolumn{1}{c|}{\textbf{1.94}} & \textbf{1.00} & \multicolumn{1}{c|}{\textbf{5.13}} & \multicolumn{1}{c|}{\textbf{3.05}} & 1.58 & \multicolumn{1}{c|}{\textbf{6.78}} & \multicolumn{1}{c|}{\textbf{4.22}} & 2.41 & \textbf{3.22} \\ \hline
    \multirow{6}{*}{MSE ($10^{-2}$)} & BF & \multicolumn{1}{c|}{3.50} & \multicolumn{1}{c|}{2.59} & 1.84 & \multicolumn{1}{c|}{3.53} & \multicolumn{1}{c|}{2.62} & 1.88 & \multicolumn{1}{c|}{3.65} & \multicolumn{1}{c|}{2.75} & 2.01 & 2.71 \\ \cline{2-12} 
     & WLOP & \multicolumn{1}{c|}{4.12} & \multicolumn{1}{c|}{2.83} & 2.08 & \multicolumn{1}{c|}{4.26} & \multicolumn{1}{c|}{2.97} & 2.03 & \multicolumn{1}{c|}{4.75} & \multicolumn{1}{c|}{3.61} & 2.42 & 3.23 \\ \cline{2-12} 
     & EC-Net & \multicolumn{1}{c|}{3.58} & \multicolumn{1}{c|}{2.54} & 1.64 & \multicolumn{1}{c|}{3.64} & \multicolumn{1}{c|}{2.65} & 1.79 & \multicolumn{1}{c|}{3.76} & \multicolumn{1}{c|}{2.82} & 2.16 & 2.73 \\ \cline{2-12} 
     & PF & \multicolumn{1}{c|}{3.46} & \multicolumn{1}{c|}{2.44} & 1.56 & \multicolumn{1}{c|}{3.54} & \multicolumn{1}{c|}{2.54} & 1.65 & \multicolumn{1}{c|}{3.73} & \multicolumn{1}{c|}{2.69} & 1.78 & 2.60 \\ \cline{2-12} 
     & RePCD & \multicolumn{1}{c|}{3.45} & \multicolumn{1}{c|}{2.43} & 1.55 & \multicolumn{1}{c|}{\textbf{3.43}} & \multicolumn{1}{c|}{\textbf{2.48}} & 1.69 & \multicolumn{1}{c|}{3.57} & \multicolumn{1}{c|}{2.69} & 1.76 & 2.56 \\ \cline{2-12} 
     & Ours & \multicolumn{1}{c|}{\textbf{3.43}} & \multicolumn{1}{c|}{\textbf{2.42}} & \textbf{1.55} & \multicolumn{1}{c|}{3.46} & \multicolumn{1}{c|}{2.50} & \textbf{1.64} & \multicolumn{1}{c|}{\textbf{3.57}} & \multicolumn{1}{c|}{\textbf{2.65}} & \textbf{1.76} & \textbf{2.55} \\ \hline
    \end{tabular}
    \caption{Quantitative comparison of various methods on synthetic models with different noise levels and point densities}
    \label{tab1}
\end{table*}

\subsection{Learnable Bilateral Filter}
\label{sub2.3}
We first establish the multi-scale neighborhoods of $\hat{p}$ as:
\begin{equation}
\footnotesize
\begin{aligned}
\hat{\wp}_{i, k} &=\left\{\hat{p}_{j,k}\mid\left\| \hat{p}_{j, k}-\hat{p}_i\right\|<r_k\right\}, \\
\wp_{i, k} &=\left\{p_{j, k}\mid\left\|p_{j, k}-\hat{p}_i\right\|<r_k\right\},
\end{aligned}
\end{equation}
where $k$ is the scale index, $\hat{p}_{j,k}, \hat{p}_i\in\hat{\mathcal{P}}$, $p_{j,k}\in\mathcal{P}$ and $r_k$ is the patch radius of the scale $k$. $\wp_{i, k}$ is the noise-free patch corresponding to the noisy version $\hat{\wp}_{i, k}$.
Before the multi-patches are fed into the parameter prediction network, they need to be preprocessed to improve the robustness of denoising. We normalize the input patches to avoid redundant degree of freedom from the observed space, i.e., $\hat{\wp}_{i, k}=$ $\left(\hat{\wp}_{i, k}-\hat{p}_i\right) / r_{\max }, \wp_{i,k}=\left(\wp_{i,k}-\hat{p}_i\right) / r_{\max }$, where $r_{\max }$ is radius of the max scale. To guarantee LBF to be invariant to rigid transformation (e.g., rotation), we rotate each patch by aligning its principle axes of PCA with the Cartesian space and obtain the rotation matrix $R$.
\begin{algorithm}[htb]
    \small
    \label{alg:lbf}
    \caption{$learnable$ $bilateral~(\hat{p}, r_k)$}
    \KwIn{Noisy point $\hat{p}\in\mathcal{\hat{P}}$, the radius $r_k$ of the $k$-th scale}
    \KwOut{Denoised point $p$}
    $\hat{\wp}_{i, k}\leftarrow$ multi-scale neighbors of $\hat{p}$\;
    $R^{-1}\leftarrow$ patches preprocessing\;
    $\sigma_d, \sigma_n = f(\hat{\wp}_{i, k})$\, // $f$ is the parameter prediction network\;
    Compute unit normal to the regression plane $n_p$ from $\hat{\wp}_{i, k}$\;
    $sum_w=0$\;
    $\delta_p=0$\;
    \For{$q\in\mathcal{N}_{r,k}$}
    {
        $d_d \leftarrow\|q-p\|$\;
        $d_n \leftarrow\left\langle q-p, n_p\right\rangle$\;
        $w=\exp -\frac{d_d^2}{2 \sigma_d^2} \exp -\frac{d_n^2}{2 \sigma_n^2}$\;
        $\delta_p=\delta_p+w d_n$\;
        $\operatorname{sum}_w=\operatorname{sum}_w+w$\;
    }
    $p \leftarrow \hat{p}+\frac{\delta_p}{\operatorname{sum}_w} n_p*{R}^{-1}$
\end{algorithm}
After preprocessing, we feed the multi-scale patches into an encoder-decode parameter prediction network, as shown in Fig. \ref{fig:overview}. The encoder module employs MLPs (64, 128, 256, 512) followed by a max-pooling operation to extract feature vectors from each patch. In this way, the multi-scale features that capture geometric information can be encoded simultaneously.
In the decoder module, we use fully connected layers (FC) to predict the parameter $\sigma_d$ for denoising and the parameter $\sigma_n$ for sharp feature preserving. To fully utilize the multi-scale information, we concatenate the multi-scale features and compute a set of weights for each patch. The final parameters predicted by the network are obtained by weighting the parameter results of all scales. Then, we compute the unit normal $n_p$ to guarantee the denoised point $p$ within a $r$-ball centered around $\hat{p}$. The displacement between the noisy point and corresponding clean point is the weighted average of the projections on the line $(\hat{p}, n_p)$ of points around $p$. Due to the input patch having been transformed into a canonical space, the final displacement should be multiplied by the inverse matrix $R^{-1}$. Finally, we get the denoised point by adding the noisy point and the displacement. The learnable bilateral filter for a given point $\hat{p}\in\mathcal{\hat{P}}$ is shown in Algorithm~\ref{alg:lbf}. 

\subsection{Loss Function}
\label{sub2.5}
The loss function is elaborately designed to achieve better denoising performance. Inspired by PF \cite{PF}, we design a geometry-aware bi-directional projection loss function to train our LBF. The loss function preserves sharp features by considering the distance and the normal similarity between the current denoising point $\hat{p}_i$ and its neighboring points of the ground-truth patch. The loss function is defined as:
\begin{equation}
\footnotesize
L_{\text {proj1}}=\frac{\sum_{p_j \in \wp_{i,k}}\left|\left(\bar{p}_i-p_j\right) \cdot n_{p_j}^T\right| \cdot \phi\left(\bar{p}_i,p_j\right) \theta\left(n_{\bar{p}_i}, n_{p_j}\right)}{\sum_{p_j \in \wp_{i,k}} \phi\left(p_i,p_j\right) \theta\left(n_{\bar{p}_i}, n_{p_j}\right)},
\end{equation}
\begin{equation}
\footnotesize
L_{\text {proj2}}=\frac{\sum_{p_j \in \wp_{i,k}}\left|\left(\bar{p}_i-p_j\right) \cdot n_{p}^T\right| \cdot \phi\left(\bar{p}_i,p_j\right) \theta\left(n_{\bar{p}_i}, n_{p_j}\right)}{\sum_{p_j \in \wp_{i,k}} \phi\left(p_i,p_j\right) \theta\left(n_{\bar{p}_i}, n_{p_j}\right)},
\end{equation}
where $\bar{p}_i$ is the filtered point of  $\hat{p}_i$, and $n_{p_j}$ (or $n_p$) is the ground-truth normal of  $p_j$ (or $p$). $\phi\left(\bar{p}_i,p_j\right)$ is a Gaussian function giving larger weights to the points near $\bar{p}_i$, defined as $\phi\left(\bar{p}_i,p_j\right)=\exp \left(-\frac{\left\|\bar{p}_i-\mathrm{p}_j\right\|^2}{\varepsilon_p^2}\right)$, where $\varepsilon_p$ is defined as $\varepsilon_p=4 \sqrt{d / m}$, where $d$ is the length of the diagonal of the bounding box of patch $\wp_{i, k}$ and $m=\left|\hat{\wp}_{i,k}\right|$. $\theta\left(n_{\bar{p}_i}, n_{p_j}\right)$ is a feature preserving function giving larger weights to the neighboring points with more similar normal to $\bar{p}_i$, defined as $\theta\left(n_{\bar{p}_i}, n_{p_j}\right)=\exp \left(-\frac{1-n_{\bar{p}_i}^T n_{p_j}}{1-\cos \left(\varepsilon_n\right)}\right)$, where $\varepsilon_n$ is the support angle ($15^{\circ}$ by default). Our bi-directional projection loss function is defined as:
\begin{equation}
    \footnotesize
    L_{\text {bi-proj}}=L_{\text {proj1}}+L_{\text {proj2}}.
\end{equation}

To make the filtered point distribution uniform, we also employ a repulsion term to penalize point aggregation. The final loss function is formulated as:
\begin{equation}
\footnotesize
L=\eta L_{\text {bi-proj }}+(1-\eta) L_{r e p}, \quad L_{r e p}=\max _{p_j \in \wp_{i,k}}\left|\bar{p}_i-p_j\right|.
\end{equation}

\begin{figure*}[t]
    \centering
    \includegraphics[width=0.88\textwidth]{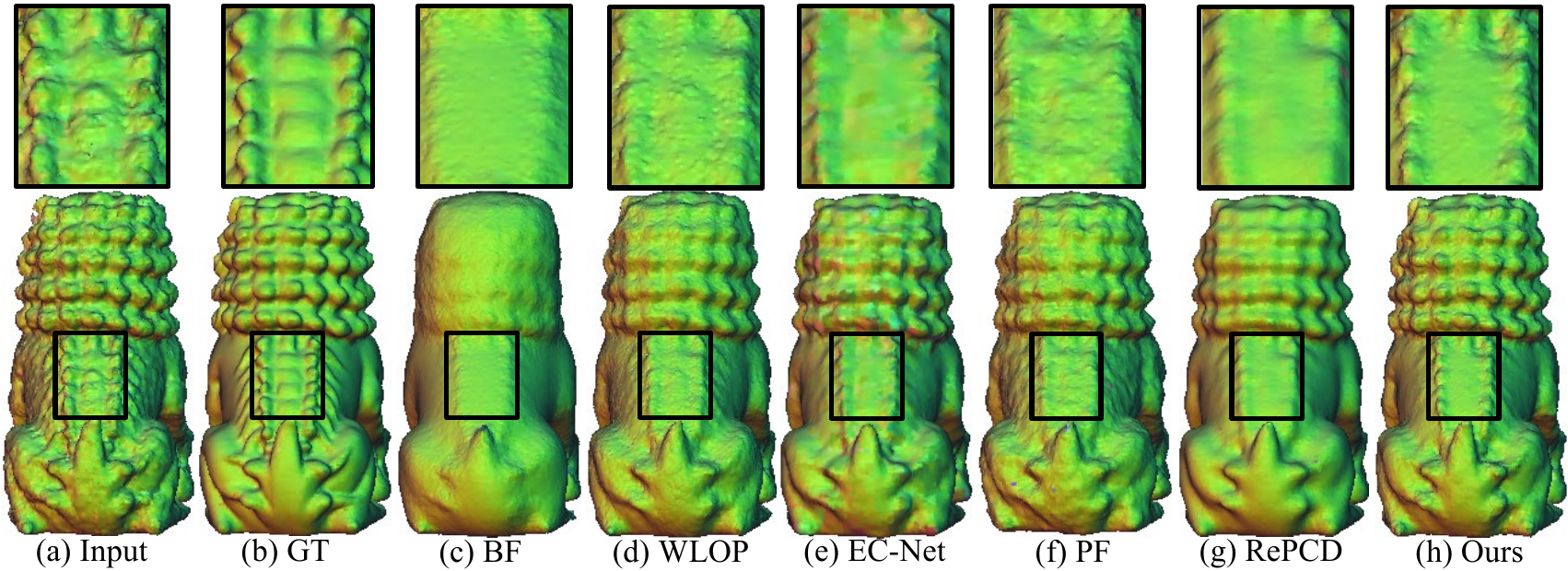}
    \caption{Visual comparisons on a synthetic model. LBF better preserves geometric features while robustly removing noise.}
    \label{fig:syn}
\end{figure*}

\begin{figure*}[t]
    \centering
    \includegraphics[width=0.88\textwidth]{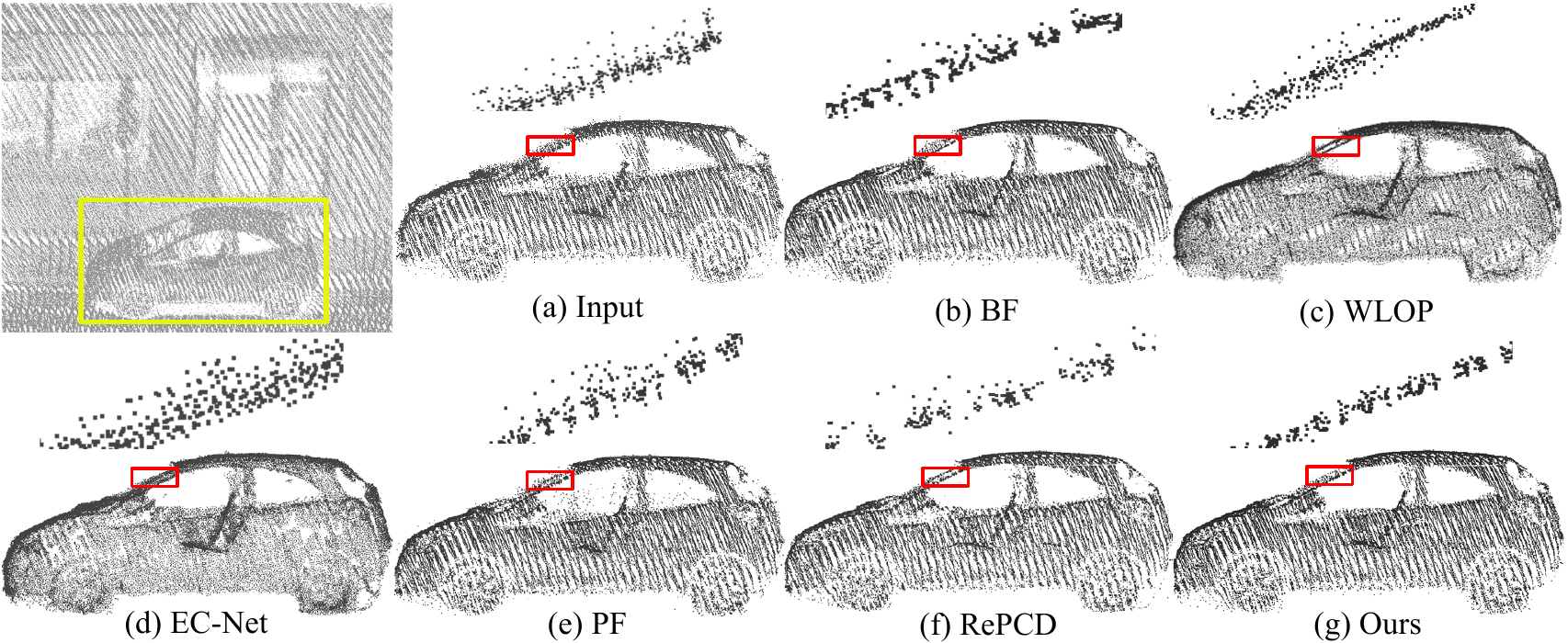}
    \caption{Visual comparisons on the real-world dataset Paris-rue-Madame. LBF behaves best to deal with outliers.}
    \label{fig:real}
\end{figure*}

\section{EXPERIMENTS}
\label{sec:experiment}
\subsection{Dataset}
\label{sub3.1}
The training dataset is provided by PF \cite{PF}, which consists of 11 CAD models and 11 non-CAD models. We randomly sample 10k, 20k, and 50k points from  each model. Then, the clean models are polluted by Gaussian noise with the standard deviations from 0.0\% to 1.5\% of the bounding box’s diagonal length. In addition to the point coordinates of each model, normal information is also required during training.

To verify the effectiveness of our method, the test dataset includes synthetic models sampled from the dataset provided by PU-GAN \cite{pu-gan} and the real-world Paris-rue-Madame dataset \cite{rue}. Similar to the training dataset, 32 synthetic models are randomly sampled to 10k, 20k, and 50k points respectively, and then contaminated by Gaussian noise with the standard deviations of 0.0\% to 1.5\% of the clean model's bounding box diagonal length.

\subsection{Implementation Details}
\label{sub3.2}
The number of scales is set as 3 and the point number of each patch $\left|\hat{p}_{i,k}\right|$ is set as 400. 
We pad the origin for patches with points less than 400 and randomly downsample for patches with points more than 400. 
The radius $r_k$ of the three patches are set to 3\%, 4\%, and 5\% of the model's bounding box diagonal length respectively. 
The whole network is trained end-to-end with a learning rate of  $1 \times 10^{-4}$ for 25 epochs. The learning rate is also decayed by 0.1 for every 5 epochs and our LBF can converge quikly. All the networks are implemented using PyTorch and trained on an NVIDIA RTX 3060 GPU.

\subsection{Results}
\label{sub3.3}
We compare our LBF with several representative point cloud denoising methods, including BF \cite{BF}, WLOP \cite{WLOP}, EC-Net \cite{EC-Net}, PF \cite{PF} and RePCD \cite{Repcd}. 
To quantitatively compare these methods, we calculate the Chamfer distance (CD) and Mean Square Error (MSE) for the synthetic noisy models in the test dataset, as reported in Table \ref{tab1}. It can be observed that the proposed LBF achieves the best results among all the competitors. 
Besides, we also present the visual comparisons on both synthetic noisy point clouds and real-world scanned point clouds. Fig. \ref{fig:syn} shows the denoised results of different methods on a synthetic model with the noise level of 0.5\% and the point density of 50k. Compared with them, LBF achieves the best denoising result while faithfully preserving detailed geometric features. Besides, our LBF suppresses heavy noise better than competitors on real-world data. For example,  Fig. \ref{fig:real} shows the denoised result of various methods on the real-world scanned dataset \cite{rue}.
Again, our method produces a more desirable result that contains no outliers.

\section{Conclusion}
\label{sec:conclusion}
Bilateral filter is validated as a robust and efficient denoising technique in point cloud denoising.
In this paper, we propose an end-to-end learnable bilateral filtering network for point cloud denoising for the first time.
Unlike the conventional BF that laboriously adjusts the same parameters for the entire point cloud, LBF uses a parameter prediction network to learn adaptive parameters for each noisy point according to its geometric characteristics. In this way, it can remove noise effectively in smooth  regions, while better preserving geometric features in sharp regions. Various experiments show the robustness of our LBF in denoising both synthetic and raw point scans. In the future, we attempt to incorporate point cloud upsampling and completion techniques into a unified point cloud consolidation framework.


\bibliographystyle{IEEEbib}
\bibliography{strings,refs}

\end{document}